\documentclass{article}
\usepackage{spconf,amsmath,graphicx}
\usepackage{latexsym}
\usepackage{multirow}
\usepackage{pgfplots}
\usepackage{amsmath,amssymb}
\usepackage{booktabs,enumitem}
\usepackage{csquotes}

\usepackage{url}
\usepackage{xcolor}
\newcommand{\ignore}[1]{}

\newcommand{\mc}{M_c}
\newcommand{\mdist}{M_{dist}}
\newcommand{\wordemb}{E^{w}}
\newcommand{\labelemb}{E^{l}}
\newcommand{\wtmat}[1]{\mathbf{w}_{#1}}

\title{Joint Learning of Word and Label Embeddings for Sequence Labelling in Spoken Language Understanding}
\name{\begin{tabular}{c}Jiewen Wu$^{1}$, Luis Fernando D'Haro$^{2}$, Nancy F. Chen$^{1}$, \\Pavitra Krishnaswamy$^{1}$, Rafael E. Banchs$^{3}$ \end{tabular}} 
\address{$^{1}$ Institute for Infocomm Research, A*STAR,  Singapore, 138632  \\ $^{2}$ Speech Technology Group, ETSIT, Universidad Polit\'{e}cnica de Madrid, 28040, Spain \\$^{3}$ Intapp Inc., 200 Portage Ave. Palo Alto, CA, 94306, USA }
\begin{document}
%
\maketitle
\begin{abstract}
We propose an architecture to jointly learn word and label embeddings for slot filling in spoken language understanding. The proposed approach encodes labels using a combination of word embeddings and straightforward word-label association from the training data. Compared to the state-of-the-art methods, our approach does not require label embeddings as part of the input and therefore lends itself nicely to a wide range of model architectures. In addition, our architecture computes contextual distances between words and labels to avoid adding contextual windows, thus reducing memory footprint. We validate the approach on established spoken dialogue datasets and show that it can achieve state-of-the-art performance with much fewer trainable parameters.
\end{abstract}

\section{Introduction}
%
%
In spoken language understanding (SLU), an essential step is to associate each word in an utterance with one semantic class label. These annotated utterances can then serve as a basis for higher level SLU tasks, such as topic identification and dialogue response generation. This process of semantic label tagging in SLU, dubbed slot filling, labels utterance sequences with tags under a specific scheme. As an example, the BIO scheme prefixes tags with one of the characters $\{$B, I, O$\}$ to indicate the continuity of a tag: Begin, Inside, or Outside, e.g., B-\emph{price} indicates this position is the beginning of the tag \emph{price}.  

Researchers also developed deep learning architecture for slot filling, e.g., \cite{Mesnil2013,Mesnil2015,inter17}. An utterance, considered to be a sequence of words, is often represented as a sequence of vectors, e.g., word embeddings or character embeddings. With encoded utterances and labels, a deep learning model then attempts to learn the associations between them. A typical sequence labelling architecture follows the RNN+CRF paradigm \cite{Reimers2017OptimalHF}, where words are processed by recurrent neural networks and the dependencies of labels are handled by the CRF layer. This kind of models are straightforward to implement and perform well, thus widely applied to sequence labelling tasks, including slot filling.

Contextual information, in addition to input words, has been widely adopted for sequence labelling. The context may refer to external knowledge  \cite{Huang2017ImprovingSF,teacher,kblstm},  dependency graph \cite{Huang2017ImprovingSF}, or sentence structures \cite{teacher}. 
Specifically, \cite{teacher} explores contextual and structural information of utterances in dialogues, while  \cite{kblstm} extends RNNs with graph embedding to learn concepts from knowledge bases and integrate the concept embedding into the state vectors of words. Simpler contextual information can also be used, for example, the context window of an input word \cite{Mesnil2013,inter17} is used explicitly to improve labeling accuracy. We believe that the explicit context windows can be omitted due to heavy memory usage. One of our objectives is to avoid using upfront context windows explicitly, while capturing such contextual information in a localized manner.

Attention-based models \cite{Bahdanau2014NeuralMT,mahovy16} are used to capture the interaction between utterances and label representation, in which the representation of semantic labels is learned and, presumably, given as part of the input to leverage the label embeddings. This is infeasible during testing as we do not know ground truth labels in advance. In \cite{news18slot}, semantic priming is proposed to learn label embeddings from word representations. In contrast,  \cite{wang_id_2018_ACL} simply uses all labels as input during test time to take advantage of label embeddings. Another approach, instead of modifying the input, develops special architectures to make use of label embeddings, similar to neural machine translation models \cite{Bahdanau2014NeuralMT,inter17}, where a predicted label is used for the subsequent prediction. 

There are several undesirable implications of the above two approaches for leveraging label embeddings. First, learning label embeddings introduces additional parameters. In case of a huge number of semantic labels, these models, sensitive to the number of trainable parameters, tend to overfit easily because spoken dialogue datasets are usually not large. Second, the proposed architecture may be too complex to be adapted for well-known paradigms, such as the straightforward RNN+CRF architecture for sequence-to-sequence learning. As an example, the proposed model in \cite{inter17} relies on the previously predicted label as contextual information to predict the next label. To adopt the idea in \cite{inter17}, the simpler RNN+CRF model requires non-trivial modifications.

%

In this paper we propose an approach to address the aforementioned weaknesses of the existing types of architecture that leverages label embeddings. To this end, we propose a model that (1) produces label embeddings without incurring many additional parameters, (2) does not need ground truth labels to be fed as input, and (3) can be integrated on top of the RNN+CRF models in a straightforward manner. Specifically, our approach calculates label embeddings from word embeddings and word-label association information from the training data. With these two inputs, label embeddings reside in the same semantic space and share the same parameters as word embeddings.  
Overall, our contributions are as follows. (1) We propose a  novel approach to jointly learn label and word embeddings, in which label embeddings are computed from word embeddings.  (2) The proposed approach is straightforward to implement as an auxiliary, without requiring labels as input. This lends itself to a wider range of sequence learning models. (3) We evaluate an RNN-based implementation of our approach, which outperforms the baseline sequence labeling model and can reach performance close to the state-of-the-art, yet with far fewer parameters and reduced memory footprint.

The organization of this paper is as follows. 
Section~\ref{sec:lbrep} elaborates the proposed model architecture. An empirical evaluation is provided in Section~\ref{sec:exp}, followed by detailed error analyses in Section~\ref{sec:analyses}. Finally, Section~\ref{sec:summary} summarizes this work and presents some future directions.
\section{Learning Label  Representation}\label{sec:lbrep}
\begin{figure}[ht!]
\centering
\includegraphics[width=0.4\textwidth,height=0.28\textheight]{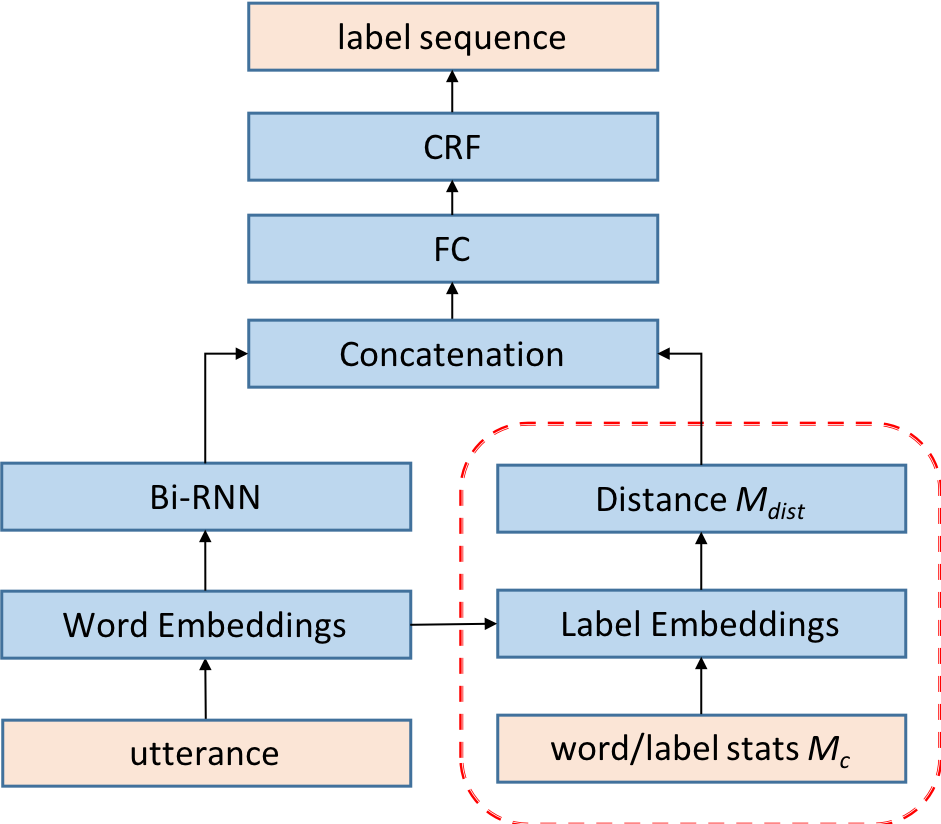}
\caption{Proposed topology. FC denotes a fully connected layer. Bi-RNN refers to bidirectional RNNs. Distance matrix $\mdist$ is passed for concatenation, providing information between words and labels.}
\label{fig:top}
\end{figure}

Assuming the utterances are encoded using word embeddings, we illustrate in the section how label embeddings can be jointly learned with word embeddings.
The overall architecture is given in Figure~\ref{fig:top}. In particular, we assume without loss of generality that an RNN+CRF neural network architecture is the baseline model for slot filling. What this paper proposes is the addition of learning label embeddings with some contextual information from the training data. The module inside the frame in Figure~\ref{fig:top} elaborates the design. The input to the framed component has two parts, one being the word embeddings encoding the utterances, the other some statistics computed between words and labels from the training data offline. Clearly, both are easily available. The output is the distances between utterances and corresponding labels. 

In what follows, we first show how label embeddings can be learned from both the word embeddings and the pre-computed statistical information. Then, we show how distances are computed between utterances and labels and how they capture local contextual information. Assume the vocabulary size for words and labels is $n$ and $m$, respectively, and the dimension of word embeddings is $d$. An utterance is a sequence of words $\langle w_1, \dots, w_i\rangle _{1\le i\le n}$, with a corresponding sequence of labels $\langle l_1, \dots, l_j\rangle _{1\le j\le m}$.

\subsection{Label Embeddings}\label{sec:lbem}
Before label embeddings are learned, we first define what statistical information needs to be computed. A straightforward method is to simply calculate the co-occurrences between words and labels in the training data.
For any label, we record, for each word, the number of times that word is tagged with this label. This results in a matrix, $\mc:m\times n$ (cf. Figure~\ref{fig:top}), that shows how every label is related to every word in the training dataset. In practice, for a label, only a few words are frequently tagged by the label, leading to a sparse matrix $\mc$. To help with distance calculation in the following steps, $\mc$ can be smoothed by adding one to all elements and normalized over the rows to allow for weighted calculations later. 

Once the word-label association matrix, $\mc: {m\times n}$, and the word embeddings $\wordemb: n\times d$ are ready,  label embeddings, $\labelemb$, can be computed as follows: 
\begin{align}
\labelemb=&(\wtmat{1}\circ\mc)(\wtmat{2}\circ\wordemb), \text{where } \nonumber
\\&\wtmat{1}: m\times 1, \wtmat{2}:n\times 1, \text{ and } (v\circ M)_{ij}=v_i M_{ij}\label{eq:labelemb}
\end{align}
In \eqref{eq:labelemb}, the learnable weights $\wtmat{1}$ and $\wtmat{2}$, are vectors, instead of matrices \cite{Hartung2017LearningCF}. This way, we can avoid incurring unnecessary parameters as only a total of $m+n$ parameters are needed. This heuristic is employed as the size of word vocabulary $n$ and/or label vocabulary $m$ can be large. 

Intuitively, the representation of a label is analogous to the centroid of a cluster of the relevant word vectors, while relevancy is defined by the pre-computed co-occurrence information.
Note that label embeddings $\labelemb$ have the same dimension as word embeddings, i.e., $d$. In addition, whether pre-trained word embeddings are used or not, label embeddings are learned in the same manner. In general, the number of parameters required to ``learn" label embeddings in Figure~\ref{fig:top} is negligible compared to using a standard embedding matrix with $n\times d$ parameters.
%
\subsection{Computing Word-Label Distances}\label{sec:lbdist}
As discussed earlier, the label representation computed by our method render a label analogous to the centroid of a cluster of  words that are likely to be tagged by the label. When a word is processed for slot filling, we can use the distances between this word and all labels for predicting the right label.

We thus use cosine similarity to calculate such distances. For a given utterances of $k$ words, a distance matrix $\mdist :k\times m$ is computed based on the word and label embeddings. Let $i, j$ denote the $i$-th word vector and the $j$-th label vector, then we can populate the distance matrix as follows:
\begin{align}
[\mdist]_{ij} = 1- \frac{\wordemb_{i}\cdot\labelemb_{j}}{\Vert \wordemb_{i}\Vert \Vert\labelemb_{j}\Vert }\label{cosine1}
\end{align} 

Note that \eqref{cosine1} computes distances between a word and all labels, which contain the label-based contextual information for every word. Conversely, word-based contextual information can be calculated for every label.  In \cite{Mesnil2015, inter17}, the context windows are used as part of the input, in contrast, the context window here is implicitly used: distances are calculated between one word and its surrounding contextual words and every label. As adapted from \cite{wang_id_2018_ACL}, for every word in the $2q+1$ context window of $i$-th word, $i-q\le p\le i+q$, we apply \eqref{cosine1} to every word vector $p$ and the $j$-th label. This is an updated version to calculate $[\mdist]^{*}_{ij}$ considering context windows, given below in \eqref{cosine2}:
\begin{align}
[\mdist]^{*}_{ij} = f_p(f_a(W_d^{\intercal}([\mdist]_{i-p, j},\ldots, [\mdist]_{i+p,j}),  \label{cosine2}
\end{align} 
where $f_p$ is a pooling function to reduce dimensions, $f_a$ an activation, $W_d$: $(2q+1)\times 1$ to be learned. 

Note that distances are computed dynamically for every utterance. Using \eqref{cosine2} necessitates more operations than using \eqref{cosine1}, but we believe that in general the rich contextual information obtained this way outweighs the increased time needed for distance computation.
%

\section{Experiments}\label{sec:exp}
To validate our approach, we implemented our proposed architecture on top of the classic RNN+CRF model, as shown in Figure~\ref{fig:top}. This section elaborates the experimental setup and compares the results with the state-of-the-art.
\subsection{Datasets}\label{sec:dataset}
We experiment with two datasets on spoken dialogues, namely, the Air Travel Information System (ATIS) task \cite{atis,atisversion} and MEDIA, French dialogues collected by ELDA \cite{media}. The statistics of the two datasets are given in Table~\ref{tb:datasets}.
\begin{table}[t!]
\begin{center}
\begin{tabular}{cccc}
\toprule
 & ATIS& MEDIA  \\\cline{2-3}\noalign{\smallskip}
\# utterances in train  &  3982 & 12908  \\
\# utterances in dev  & 995  & 1259 \\
\# utterances in test  & 893 & 3005 \\
\# labels  & 127  & 138 \\
vocab. size & 572 & 2427\\
max utterance length & 46  &  192  \\
\bottomrule
\end{tabular}
\end{center}
\caption{Statistics of datasets. The split of datasets were used as is.}
\label{tb:datasets}
\end{table}
For both ATIS and MEDIA, entities are used as the utterance input. In contrast to \cite{inter17}, no context windows were used as part of the inputs in our models. Instead, contextual information has been exploited at different stages by our models, as described in Section~\ref{sec:lbrep}. 

Note that there are many short utterances in MEDIA; in particular, MEDIA has over 4,000 utterances consisting of a single word ($\sim$30\% total utterances). 


\subsection{Setup and Hyperparameters}\label{sec:hyper}
%
For all experiments, we used a set of fixed hyperparameters for  comparison. The dimension of word embedding in all datasets is 300. For recurrent neural networks, we used GRU, with a recurrent dropout of 0.5 between recurrent units \cite{Reimers2017OptimalHF}. The Bi-GRU (cf. Figure~\ref{fig:top}) has 60 units. For the weights in \eqref{eq:labelemb}, a regularization of $10^{-6}$ was also imposed. For \eqref{cosine2} we chose $5$ as the the size of context window and used max-pooling and ReLU for $f_p$ and $f_a$ respectively. The stride for max-pooling was set to $10$.
%
%
During the learning phase, a mini-batch size of 32 and an initial learning rate of 0.004 was used with the Nadam optimizer \cite{dozat2016incorporating} to minimize the cross-entropy loss. The learning rate was reduced by 50\% when no improvement was seen after three epochs. 


\subsection{Results}\label{sec:res}
In this section the CoNLL-F1 scores are reported in Table\ref{tb:resfull}. The experiments were run on an NVIDIA Tesla V100 16GB GPU, and the F1 scores are computed from three independent runs, capped at 30 epochs.
In this section, our proposed architecture in Figure~\ref{fig:top} is referred to as \textbf{LE}, and the architecture without the modules in the framed box is considered to be the baseline \textbf{BL}. We also denote by \textbf{SOTA} the state-of-the-art approach for learning label embeddings in \cite{inter17}.  

\noindent\textbf{Reduced MEDIA.} We also investigated the impact of data reduction in MEDIA.
For reduction, we rank vocabulary by word frequency in the training and development sets. For each word we choose the first $m$ utterances containing the word, and update the vocabulary coverage after each selection of utterances. Once the full vocabulary is covered, a reduced dataset is obtained. By varying the number $m$, reduced datasets with different sizes can be obtained. Note that in all reduced datasets, the vocabulary  remain the same size. Our objective is to verify the robustness of our approach when only (reasonably) limited data is available. 

\noindent\textbf{Baseline BL.}  Given that ATIS is considered simple for slot-filling \cite{Mesnil2015,inter17}, we focus on the results for MEDIA. Note that the baseline approach in \cite{inter17}, also Bi-GRU+CRF, achieved an F1 score of 86.69 for MEDIA, which is higher than our baseline (86.2).  
We chose our own implementation (Bi-GRU+CRF) as the baseline because the baseline in  \cite{inter17} requires explicit use of context windows as input, leading to significant memory footprint. Also, our baseline has only 668k trainable parameter, compared to 2.3 million in the baseline of  \cite{inter17}, about 70\% reduction in parameters. We could, in principle, increase the number of units of GRUs and the fully-connected layers to achieve a higher F1 score, but our work is to show how label embeddings help build comparable models with far \emph{fewer parameters} and \emph{less memory footprint}.

\begin{table}[th!]
\begin{center}
\begin{tabular}{cccc}
\toprule
 & \textbf{BL}&\textbf{LE} & \textbf{SOTA} \\
 \cline{2-4}
 \noalign{\smallskip}
\textbf{A} & 95.33 (0.21)  &95.21 (0.15) & 95.74 (0.02)\\
\textbf{M}  & 86.2 (0.22) & 86.78 (0.02)& 86.97 (0.12) \\
\textbf{M}$_{5}$& 84.68 (0.02)  &   84.65 (0.04)& --\\
\textbf{M}$_2$&   82.97 (0.03)  &   83.53 (0.09) & --\\
\noalign{\smallskip}\cline{2-4}\noalign{\smallskip}
\# para. \textbf{A}  & 334K & 346K & 340K \\
\# para. \textbf{M} & 668K & 682K & 1743K \\
\bottomrule
\end{tabular}
\end{center}
\caption{(conll-)F1 of the two datasets. \textbf{A} (\textbf{M} resp.) refers to the ATIS (MEDIA resp.) dataset.
\textbf{M}$_{5}$ is the reduction of MEDIA with $m=5$ as discussed in Section~\ref{sec:dataset}. Similarly, \textbf{M}$_2$ is the reduction of MEDIA with $m=2$. \textbf{M}$_{5}$ has  3117/622 utterances in the training/development set, while \textbf{M}$_2$ has  2010/463 utterances. In all cases, the test set is not reduced. }
\label{tb:resfull}
\end{table}


Table~\ref{tb:resfull} shows the results of our proposed architecture on both ATIS and MEDIA. For ATIS, we did not see significant difference between \textbf{BL} and \textbf{LE} models, both of which, however, are lower than 95.74\% by \textbf{SOTA}. However, since ATIS is considered to be a simple dataset for slot filling, the differences are not significant. For MEDIA, our approach \textbf{LE} obtained a significantly higher F1 compared to the baseline \textbf{BL}. The difference between \textbf{SOTA} and \textbf{LE} is also small. Furthermore, note that our model \textbf{LE} seemed to be very stable, with very low variance across runs. 

Table~\ref{tb:resfull} also include various MEDIA versions reduced by more than 70\% in size. In the reduced version, the number of utterances that consist of a single word were significantly lowered. Though our approach was not better than the baseline for \textbf{M}$_{5}$, it  outperformed the baseline by 0.5\% for \textbf{M}$_{2}$. Together with the results for the full MEDIA dataset, we can see that our proposed architecture seems to be more stable and robust than the baseline approach. 

In Table~\ref{tb:resfull} we also listed the number of trainable parameters in each model. Note again that our word and label embeddings have 200 dimensions in both ATIS and MEDIA, while \cite{inter17} used 100 and 200 dimensions for ATIS and MEDIA, respectively. Even with much fewer dimensions, the Jordan network based model in \cite{inter17} still requires more than 1.7 million parameters, while, in comparison, our model needs only 682,000 parameters and can achieve comparable performance. There are at least two reasons why our approach requires far fewer parameters. (1) The approach presented in \cite{inter17} requires context windows of words as explicit input, which can significantly increase the subsequent layers' parameter space. Instead, our approach uses a few optimizations to internalize the contextual information at different stages (cf. the framed box in Figure~\ref{fig:top}). (2) Our approach learns label embeddings from word embeddings and training data statistics, so the parameters are minimized as only the majority of parameters are attributed to word embeddings alone. This can be seen from Table~\ref{tb:resfull}: \textbf{LE} adds only about 2\% parameters compared to the baseline approach \textbf{BL}.

Next, we investigate further the impact of different layers introduced in our approach. Table~\ref{tb:opt} shows the ablation study on different optimizations introduced in the architecture. 
\begin{table}[th!]
\begin{center}
\begin{tabular}{lc}
\toprule
 & F1 \\
 \cline{2-2}\noalign{\smallskip}
Full & 86.78\\
(a) Use Equation \eqref{cosine1} instead of \eqref{cosine2}  & 86.26  \\
(b) Use attention instead of concatenation & 86.25
\\\bottomrule \noalign{\smallskip}
\end{tabular}
\label{tb:opt}
\caption{Optimization for improving F1 in MEDIA.}
\end{center}
\end{table}
We found that 
leveraging contextual information to compute distances instead of using the word alone is more beneficial, as shown by (a). 
Similarly, it turned out that attentive models does not work well in this model architecture for MEDIA. Instead, concatenation of distances as residual information with the base models enhances the performance.

\section{Error Analyses}\label{sec:analyses}
We have shown in the previous section that our proposed method (\textbf{LE}) reaches SOTA results in terms of (conll-)F1 score. In this section, we focus on analyzing the errors and successes of \textbf{LE} in comparison with the baseline (\textbf{BL}) over the MEDIA test dataset. For similar analyses on the ATIS dataset, interested readers can refer to \cite{bechet2018atis}. 

\subsection{Output Differences}
This section presents two types of analyses on the output. The first is to check whether the two models, \textbf{LE}  and \textbf{BL}, share similarity in the output before CRF. The second performs detailed analyses over the final output. 
\subsubsection{Output before CRF}
Consider the architecture presented in Figure~\ref{fig:top}. We show a comparison of the output of the fully-connected layer right before CRF between \textbf{LE}  and \textbf{BL}. This layer resembles the attention mechanism, where word features are learned against label features. Since CRF is used to capture the label dependencies, we investigate the raw output (from the fully-connected layer) before CRF to study whether the two models agree on the word to label mapping. 

Specifically, when a test instance is evaluated, the output of the FC layer is l2-normalized. The normalized output can be thought of as the attention weights over all labels. The weights are accumulated for all test instances and then normalized again to simulate the global attention over all labels. Thus, we obtain a matrix with dimension $2427\times 138$ as there are 2427 words and 138 labels in MEDIA. Note that, however, many words do not appear in the test set. 
For this matrix, we perform word-level, paired Wilcoxon signed-rank tests, i.e., for a word that does appear in the test set, one sample is obtained from the fully connected layer, FC, output of \textbf{LE}, the other sample from the FC output of \textbf{BL}. The null hypothesis is that the two paired samples come from the same distribution. Our hypothesis testing shows that, among the 1218 words appearing in the test dataset, 157 words have a sufficiently small p-value ($\le$0.05) to reject the null hypothesis. That is, most words (87.1\%) probably have their two models' FC output from the same distribution, while the FC output of 12.9\% words probably follow different distributions in two models. We notice that some of the latter kinds of words are inflected words, for instance, \emph{adultes}.

Based on the above tests, both \textbf{LE} and \textbf{BL} agree on the learning of most of the word-to-label over labels, yet they also exhibit differences in terms of the word-to-label weight distribution. Next, we discuss in detail the actual output (i.e., the CRF output) differences between the two models.

\subsubsection{Final Output Errors}
We first show errors by two levels in Table \ref{tb:baseline_final_errors}, i.e., errors in utterances and words. In the test dataset of MEDIA, there are 3,005 utterances and 25,997 words in total. 
In the table we distinguish between errors when BIO tags are considered and when they are not. In the latter case, the idea is to show whether the system can predict the ``concept" correctly, while ignoring the BI prefixes. 
\begin{table*}[th!]
\centering
\begin{tabular}{ccccc}
\toprule
\multicolumn{1}{l}{\multirow{2}{*}{\textbf{}}} & \multicolumn{2}{c}{\textbf{With BIO Tags}} & \multicolumn{2}{c}{\textbf{Without BIO Tags}} \\ \cline{2-5} 
             & \textbf{BL}     & \textbf{LE}     & \textbf{BL}       & \textbf{LE}      \\  
\textbf{Words with errors}                       & 2877                  & 2744   (-0.5\%)             & 2388                    & 2309  (-0.3\%)                \\  
\textbf{Utterances with errors}                   & 745                   & 708 (-1.2\%)               & 732                     & 700    (-1\%)                \\ 
\textbf{Utterances with errors (shared+unique)} 
 & 632+113 & 632+76 & 621+111 & 621+79\\
\bottomrule
\end{tabular}
\caption{Comparing errors between baseline (\textbf{BL}) and our (\textbf{LE}) model.}
\label{tb:baseline_final_errors}
\end{table*}

On the utterance level, our model in both cases (with or without BIO prefixes) results in fewer errors. Even though the overall F1 score of our model is only $0.58\%$ better than the baseline method, our utterance-level error rate has been doubly reduced. Without considering the BIO tags, similar conclusions can be drawn. The analyses also suggest that our method work better with CRF when BIO-tags are considered. 

At the word level, we observe \textbf{LE} is able to reduce up to 0.5\% compared to \textbf{BL}. Compared with the utterance level, word errors are probably localized to specific utterances in MEDIA: a high word level error rate does not necessarily lead to a high utterance level error rate. 
On the other hand, if we consider predicting the labels without BIO tags, we can reduce the word error rate significantly, from 2,744 to 2,309 (a 15.8\% relative reduction) for \textbf{LE}, and from 2,877 to 2,388 (a 17.0\% relative reduction) for \textbf{BL}.


\begin{table}[bh]
\centering
\begin{tabular}{ccc}
\toprule
\textbf{Words}   & \textbf{BL} & \textbf{LE} \\ 
\cline{2-3}
chambre / room   & 39        &  34        \\ 
h\^otel / hotel    & 36                               & 28                               \\  
trois / three    & 23                               & 14                               \\ 
prix / price     & 20                               & 18                               \\ 
chambres / rooms & 15                               & 19                               \\ 
\bottomrule
\end{tabular}
\caption{Top-5 mislabelled (non-stop) words by \textbf{BL} and \textbf{LE}.}
\label{tb:common_mislabelled_comparison}
\end{table}

We show in Table~\ref{tb:common_mislabelled_comparison} some example words in MEDIA that are mislabelled for slot filling. 
%
%
From the results, \textbf{LE} make much fewer errors than \textbf{BL} for most words, except for the word \emph{chambres}. Table~\ref{tb:common_mislabelled_comparison} does not consider stop words, e.g., determinants, prepositions. When stop words are considered, for words mislabeled at least 15 times, both \textbf{BL} and \textbf{LE} show a similar number of errors. However, some of the errors might be due to annotation errors (discussed later). Among other words, there are about 10\% fewer errors made by \textbf{LE} model than by \textbf{BL}. 

\subsection{Annotation Errors}
In order to analyze the annotation errors, we sampled utterances with errors and selected 120 of them. We found that 25\% of the word errors were independent from the systems' performance and those errors fall into two major categories: errors in the provided ground truth data and inconsistency across annotations.

\subsubsection{Ground Truth Errors} When a word is incorrectly labeled by the original annotator (i.e. tagging errors in the ground truth). In this case, even though the predictions may be correct, they are still considered to be wrong when compared with the ground truth. We found that around 8.0\% of the word errors in the 120 utterances fall in this category. For instance, words such as \emph{euh}, \emph{hum}, \emph{ben} and \emph{ah} are French interjections, which should be always labeled \textbf{O}. However, many of them in the ground truth were labeled using the surrounding label in the sentence. An example is given below.

\begin{itemize}[leftmargin=*,topsep=0pt,itemsep=0ex,partopsep=1ex,parsep=1ex]
\item \textbf{Utterance}: deux, cent, vingt,  six, euh  (two hundred and twenty six euh)
\item \textbf{Ground truth}: B-nombre, I-nombre, I-nombre,  I-nombre, I-nombre
\item \textbf{Prediction}: B-nombre, I-nombre, I-nombre,  I-nombre, O
\end{itemize}	

As we can see, the word \emph{euh} is semantically meaningless in the utterance, but the annotation assigns it with the previous semantic label \emph{nombre}. However, the predicted label O seems to be more reasonable.

\subsubsection{Ground Truth Inconsistencies} Even though the same word has the same semantic meaning in two different sentences, it is labeled differently. This type of inconsistency can confuse the models during training. Although our analyses were carried out over the test set, we believe such inconsistencies exist in the much large training set. In the 120 utterances, around 17.0\% of the word errors exhibit inconsistencies. As an example, both of the following utterances contain the word \emph{dans} (meaning \emph{in} in English). In the first utterance, the word  is labelled as the beginning of a class (prefixed B), while in the other sentence it is labeled irrelevant (O for outside). 

\begin{itemize}[leftmargin=*,topsep=0pt,itemsep=0ex,partopsep=1ex,parsep=1ex]
\item \textbf{Utterance 1}: une(B), semaine(B), de(O), vacances(O), \textbf{dans(B)}, un(B),  h\^otel(B)  (a holiday week in a hotel) 
\item \textbf{Utterance 2}: avec(B), parking(I),  priv\'e(I),  \textbf{dans(O)}, l'(B), h\^otel(I) (with private parking in the hotel)
\end{itemize}
In fact the word \emph{dans} has the same semantic meaning in both utterances, but the ground truth labels are inconsistent. 

\subsection{Modeling Errors}
We further analyze the actual performance of our proposed \textbf{LE} system. We found that 15.8\% of the word errors are attributed to the incorrect prediction of BIO tags, but the semantic classes are correctly predicted. For instance, ``B-r\'eponse" is predicted instead of ``I-r\'eponse". A common scenario is that the model labels a subset of words correctly out of all words, so the BI-labels are shifted accordingly. As an example, the utterance ``\emph{voil\`a, ok}" has the correct labels ``\emph{B-r\'eponse, I-r\'eponse}", but the model predicts ``\emph{O, B-r\'eponse}." Here, although the model understands the second word is a response, the predictions commit two errors instead of one.
On the other hand, we found that over 57\% of the word errors involve the label ``O". It is difficult to conclude the percentage of human annotation errors or modeling errors, since most of these errors occur on stop words.


\section{Conclusions and Future Work}\label{sec:summary}
We have demonstrated an approach to leverage word embeddings for computing label embeddings by observing that in natural language, the context of a word can indicate the classes the next word(s) should belong to. In addition to word embeddings, we also exploit statistical information available in the training data about the word-label association. Such association information provides a basis to gauge how labels are connected with words. Compared to most existing approaches for learning label embeddings, our approach does not require ground truth labels as input and can be applied on top of the general RNN+CRF paradigm used for sequence-to-sequence learning. 

Our results suggest that the approach can achieve comparable performance with the state-of-the-art on the dataset MEDIA, and outperform the a baseline that does not learn label embeddings. In particular, our model needs only 40\% of the total number of parameters used by the state-of-the-art label embedding approach, while achieving similar performance. In addition, our model only adds a small amount of parameters to the baseline model. Also, the word-label co-occurrences can be computed offline and used in both training and deployment stages.  

%

There are a number of future extensions that we plan to explore. First, there is a time-space trade off using our proposed method.  As the number of cosine distances in Eq. (3) depends on the number of the words in each utterances, the number of labels, and the context window size, the time needed for each epoch will increase by several folds. We plan to further reduce the number of distance computation in Eq. (3) as much as possible.
Second, we plan to experiment with other datasets designed for relevant sequence labelling tasks, such as semantic role labelling and named entity recognition. The purpose is to confirm if our proposed label embeddings can scale to label sets in various sizes. Finally, we will perform more hyper-parameter optimization \cite{bergstra2013hyperopt}. 

\section{Acknowledgements}
This research is supported by the Agency for Science, Technology and Research (A*STAR) under its AME Programmatic Funding Scheme (Project \#A18A2b0046) and the Spanish AMIC project (TIN2017-85854-C4-4-R). We thank Ignacio Gonz\'alez Godino (Universidad Polit\'ecnica de Madrid) for performing the error analyses and related experiments. 

\vfill
\pagebreak
\bibliographystyle{IEEEbib}
\bibliography{media}
\end{document}